\documentclass{bmvc2k}

\usepackage{graphicx}
\usepackage{comment}
\usepackage{amsmath,amssymb} 
\usepackage[
  separate-uncertainty = true,
  multi-part-units = repeat
]{siunitx}
\usepackage{tikz}
\usepackage{color}
\usepackage{multirow}
\usepackage{multicol}
\usepackage{lipsum}
\usepackage{graphicx, caption}
\usepackage{newtxtext,newtxmath}
\usepackage{newtxmath}
\usepackage{algorithm}
\usepackage{algpseudocode}
\usepackage{varwidth}
\usepackage{tabulary,booktabs}

\title{N2NSkip: Learning Highly Sparse Networks using Neuron-to-Neuron Skip Connections}

\addauthor{Arvind Subramaniam}{arvind1096@gmail.com}{1}
\addauthor{Avinash Sharma}{asharma@iiit.ac.in}{2}

\addinstitution{
 International Institute of Information Technology\\
 Hyderabad, India
}

\runninghead{Subramaniam, Sharma}{Learning Sparse Networks using N2NSkip Connections}

\begin{document}

\maketitle

\begin{abstract}
The over-parametrized nature of Deep Neural Networks (DNNs) leads to considerable hindrances during deployment on low-end devices with time and space constraints. 
Network pruning strategies that sparsify DNNs using iterative prune-train schemes are often computationally expensive. As a result, techniques that prune at initialization, prior to training, have become increasingly popular. In this work, we propose neuron-to-neuron skip (N2NSkip) connections, which act as sparse weighted skip connections, to enhance the overall connectivity of pruned DNNs. 
Following a preliminary pruning step, N2NSkip connections are randomly added between individual neurons/channels of the pruned network, while maintaining the overall sparsity of the network.
We demonstrate that introducing N2NSkip connections in pruned networks enables significantly superior performance, especially at high sparsity levels, as compared to pruned networks without N2NSkip connections.  
Additionally, we present a heat diffusion-based connectivity analysis to quantitatively determine the connectivity of the pruned network with respect to the reference network.
We evaluate the efficacy of our approach on two different preliminary pruning methods which prune at initialization, and consistently obtain superior performance by exploiting the enhanced connectivity resulting from N2NSkip connections. 

\end{abstract}

\section{Introduction}

Classical Deep Neural Networks (DNNs) are primarily motivated from the connectionist approach to cognitive architecture - a paradigm that proposes that the \textit{nature of wiring} in a network is critical for the development of intelligent systems \cite{fodor1988connectionism,holyoak1987parallel}. However, despite bringing about significant advances in tasks such as cognitive planning, speech processing, object detection, their over parametrized nature, as well as high memory requirements, result in increasingly complex training and inference procedures. 
Initial efforts to mitigate this issue primarily applied iterative connection sensitivity-based pruning, yielding sparse networks with a marginal drop in performance \citep{lecun1990optimal,hassibi1993optimal}. Over the years, with the advent of deep learning, there has been a renewed interest in iterative network sparsification \cite{han2015learning,liu2017learning,dettmers2019sparse,mostafa2019parameter}. However, owing to the computationally expensive nature of iterative pruning strategies, methods that prune the network at initialization are often desirable.

\begin{figure*}[!t]
\begin{center}
\scalebox{0.70}{
\includegraphics[scale=0.29]{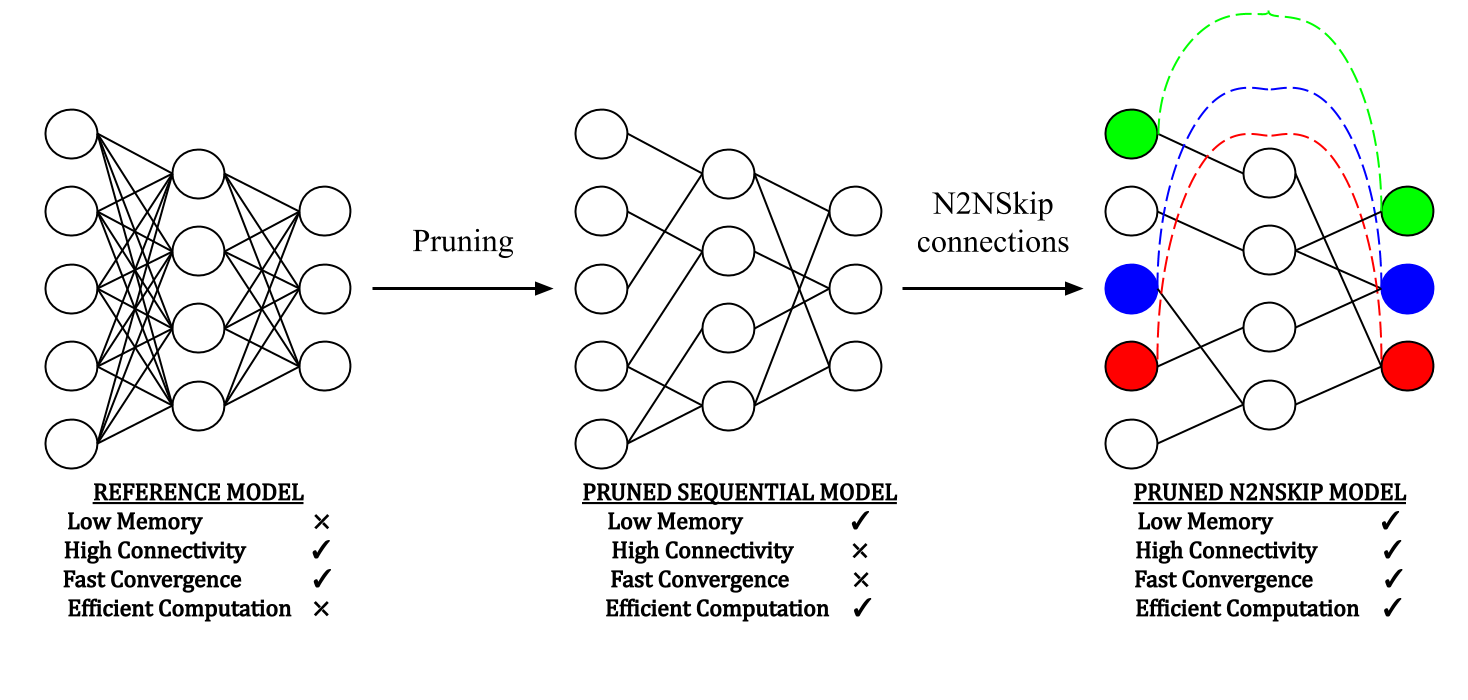}
}
\end{center}
\caption{After a preliminary pruning step, N2NSkip connections are added to the pruned network while maintaining the overall sparsity of the network. N2NSkip connections to a layer can alternatively be visualized as sparse weighted skip connections shown in Fig. ~\ref{fig:N2NSkip}}
\label{fig:neuron}
\end{figure*}

\noindent
Recently, instigated by the lottery ticket hypothesis \cite{frankle2018lottery}, there has been an increased interest in pruning networks at initialization \cite{prabhu2018deep,lee2018snip}. 
However, methods that prune at initialization lead to pruned networks that often suffer from relatively slow convergence rates as compared to the reference network. Owing to the high sparsity of the pruned network, the decrease in overall gradient flow results in inferior connectivity as compared to the reference network. 
Hence, we ask the question: Is it possible to prune a network at initialization (prior to training) while maintaining rich connectivity, and also ensure faster convergence? 

\noindent
We attempt to answer this question by emulating the pattern of neural connections in the brain.
Cognitive science has shown that neural connections in~the brain are not purely sequential but composed of a large number of skip connections as well \cite{fitzpatrick1996functional,thomson2010neocortical}. The lognormal distribution connectivity demonstrated by \cite{oh2014mesoscale} suggests the presence of sparse, long-range neuron-to-neuron connections in the brain. Although the concept of skip connections is well-established \cite{he2016deep},  these skip connections are primarily dense activations and have not been associated with learnable parameters.  

\begin{figure}[!t]
\begin{center}
\begin{tabular}{cc}
\scalebox{0.8}{
\includegraphics[width=5.2cm, height=3.9cm]{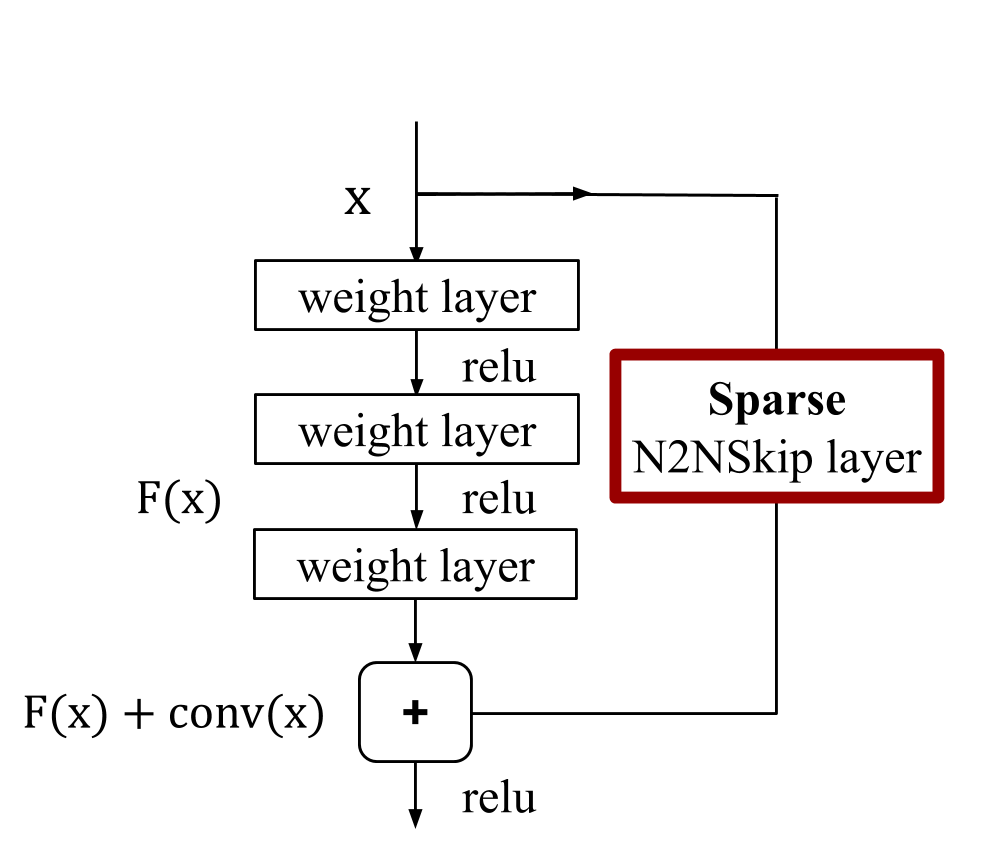}} &
\scalebox{0.8}{
    \includegraphics[width=5.2cm, height=3.9cm]{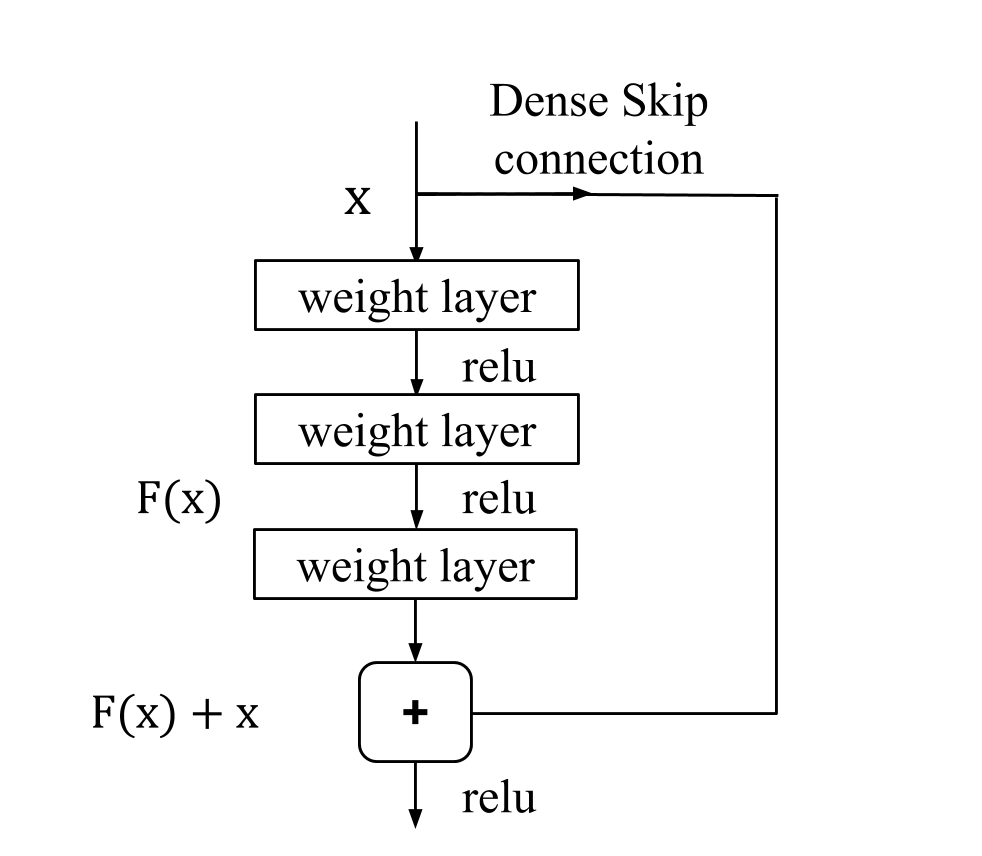}}\\
(a) N2NSkip connections & (b) Conventional skip connections\\
\vspace{1mm}    
\end{tabular}
\end{center}
\caption{As opposed to conventional skip connections, N2NSkip connections act as skip connections between  non-consecutive layers of the network, and are parametrized by sparse learnable weights.}
\label{fig:N2NSkip}
\end{figure}

\noindent
In this paper, inspired by the pattern of skip connections in the brain, we propose \textit{sparse, learnable} neuron-to-neuron skip (N2NSkip) connections, which enable faster convergence and superior effective connectivity by improving the overall gradient flow in the pruned network. N2NSkip connections regulate overall gradient flow by learning the relative importance of each gradient signal, which is propagated across non-consecutive layers, thereby enabling efficient training of networks pruned at initialization (prior to training). This is in contrast with conventional skip connections, where gradient signals are merely propagated to previous layers.  
We explore the robustness and generalizability of N2NSkip connections to different preliminary pruning methods 
and consistently achieve superior test accuracy and higher overall connectivity. A formal representation of N2NSkip connections is illustrated in Fig.~\ref{fig:N2NSkip}, where the proposed N2NSkip connections can be visualized as sparse convolutional layers, as opposed to ResNet-like skip connections. Fig.~\ref{fig:neuron} provides a detailed visualization of how a sparse N2NSkip layer is constructed.    
\noindent
Additionally, our work also explores the concept of connectivity in deep neural networks through the lens of heat diffusion in undirected acyclic graphs \cite{thanou2017learning,kondor2002diffusion,sharma2012representation}. We propose to quantitatively measure and compare the relative connectivity of pruned networks with respect to the reference network by computing the Frobenius norm of their heat diffusion signatures at saturation. 
These heat diffusion signatures are obtained by first modeling pruned (and subsequently trained) networks as weighted undirected graphs followed by computation of their saturated heat distribution vector. By comparing the difference in heat signatures of two networks, we hope to establish a strong correlation between network performance accuracy and their effective connectivity.
Our key contributions are as follows:

\begin{itemize}
  \item \textit We propose N2NSkip connections which significantly improve the effective connectivity and test performance of sparse networks across different datasets and network architectures (Section ~\ref{sec:N2NSkip}). Notably, we demonstrate the generalizability of N2NSkip connections to different preliminary pruning methods and consistently obtain superior test performance and enhanced overall connectivity (Section ~\ref{sec:exp}).  
  \item We propose a heat diffusion-based connectivity measure to compare the overall connectivity of pruned networks with respect to the reference network (Section ~\ref{sec:connectivity}). To the best of our knowledge, this is the first attempt at modeling connectivity in DNNs through the principle of heat diffusion.     
  \item We empirically demonstrate that N2NSkip connections significantly lower performance degradation as compared to conventional skip connections, resulting in consistently superior test performance at high compression ratios. (Section ~\ref{sec:resskip}). 
\end{itemize}

\section{Related Work}


\label{gen_inst}
Previous approaches to network pruning primarily fall under four categories: pruning after training (train-prune), pruning before training (prune-train), dynamic pruning and pruning during training. These methods are further elaborated below.

\noindent
\textbf{Pruning after Training.} 
Pruning is performed on pretrained models by eliminating redundant weights, followed by a finetuning process to obtain a sparse network that least degrades the overall performance. 
Two popular approaches to identify redundant weights in pretrained models have been magnitude-based pruning \cite{han2015deep,han2015learning} and Hessian-based pruning \cite{lecun1990optimal,hassibi1993optimal}. While the former directly removes weights with magnitude lesser than a specified threshold, hessian-based methods use second derivative information to compute saliencies for each parameter, and eliminate weights with lower saliency. 
As a result, connections are removed based on how they affect the loss, as opposed to magnitude-based pruning, in which important weights may be unintentionally removed.   

\noindent
\textbf{Pruning during Training.} Pruning is performed iteratively throughout training, by reducing the number of redundant weights during every iteration of training. Most methods sparsify the network by eliminating weights below a certain threshold, and propose to gradually increase the sparsity of the network
\cite{liu2017learning,dettmers2019sparse,mostafa2019parameter}. Since the criteria to prune the weights solely depends on the magnitude of the weights, these approaches usually involve multiple prune-train cycles and are computationally expensive to deploy.


\noindent
\textbf{Pruning before Training.} Methods which focus on structured simplification such as low-rank approximation, filter pruning, pruning using expander graphs and matrix factorization have also been proposed \cite{prabhu2018deep}.  The Lottery Ticket Hypothesis \cite{frankle2018lottery} employed an iterative magnitude-based pruning method to find sub-networks within a dense network, following which sub-networks are re-initialized and trained in the standard manner. Recently, a single-shot prune-train approach based on connection sensitivity was proposed, in which the importance of each weight is determined by the effect of multiplicative weight perturbations on the overall loss of the network \cite{lee2018snip}.

\section{Method}
\subsection{Neuron-to-Neuron Skip Connections}
\label{sec:N2NSkip}

\noindent
The basic building block consisting of an N2NSkip connection, shown in Fig.~\ref{fig:N2NSkip}, can be parametrized by $\{W_{j} \in \mathbb{R}^{C_{j+1} \times C_j \times f \times f}, 1 \leq j \leq 3 \}$, where $W_{j}$ denotes the weight matrix in the $j$-th layer and $C_j$ denotes the number of input channels in convolutional layer $j$.

For a given sparsity, neurons in layer $l$ are randomly connected to neurons in layer $l+k$, while maintaining the overall sparsity of the network. This is in contrast to skip connections employed in ResNet, where the dense output activation of a layer $l$ is merely added to the output of the layer $l+k$.
Given weight matrices of size $C_1 \times C_2 \times f \times f, C_2 \times C_3 \times f \times f$  and $C_3 \times C_4 \times f \times f$, the weight matrix in the N2NSkip layer is a sparse matrix of dimension $C_1 \times C_4 \times f \times f$.      
Eq. \ref{eq1} and Eq. \ref{eq2} explain the difference between conventional skip connections and the proposed N2NSkip connections. 
\begin{equation}
a^{l+k} = g(z^{l+k} + a^l).
\label{eq1}
\end{equation}
\begin{equation}
a^{l+k} = g(z^{l+k} + g(conv(a^l)).
\label{eq2}
\end{equation}
Here, \(a^l\) refers to the output activation of layer \(l\), \(z^l\) denotes the output of layer \(l\) prior to the non-linearity, $k$ refers to the number of sequential layers skipped and \(g\) is the nonlinear activation.  In Eq. \ref{eq2}, \(conv\) refers to a sparse convolutional operation on the output activation of layer \(l\). Despite adding another layer, the overall sparsity of the network is maintained. 
In other words, a pruned sequential network having a density of $10\%$ is transformed into a pruned N2NSkip network having $5\%$ skip connections and $5\%$ sequential connections.
Notably, we demonstrate the generalizability of N2NSkip connections to different preliminary pruning methods given below: 
\begin{enumerate}
\item \textit{Randomized Pruning (RP)} - We apply random sparsification to the reference network \cite{prabhu2018deep}, in which weights are randomly pruned at initialization, prior to training, while maintaining rich connectivity throughout the network (represented by \textit{RP}). N2NSkip connections are then added to the pruned network while maintaining the same sparsity. We represent these pruned networks with N2NSkip connections by \textit{N2NSkip-RP}.   
  \item \textit{Connection Sensitivity Pruning (CSP)} - We apply connection sensitivity-based pruning to the reference network \cite{lee2018snip}, where weights having minimal impact on the overall loss are deemed redundant and removed (represented by {\textit{CSP}}). N2NSkip connections are then added to the pruned network while maintaining the same sparsity. We represent these pruned networks with N2NSkip connections by \textit{N2NSkip-CSP}.
\end{enumerate}

\begin{figure}[!t]

\includegraphics[scale=0.36]{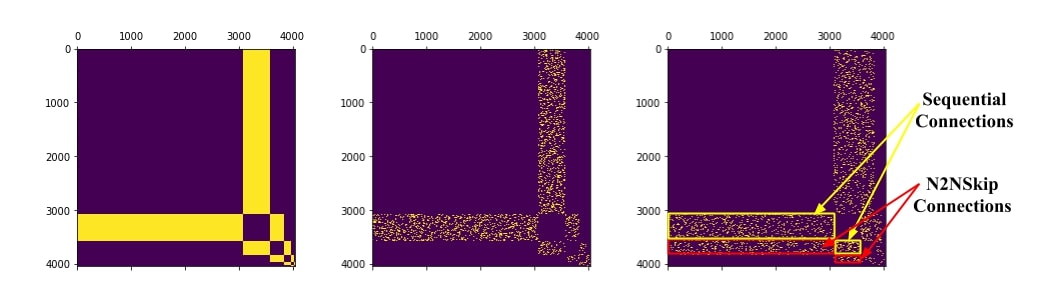}
\caption{Binary adjacency matrices for (a) Reference network (MLP) (b) Pruned network at a compression of 5x (randomized pruning) (c) N2NSkip network at a compression of 5x (10\% N2NSkip connections + 10\% sequential connections). Considering each network as an undirected graph, we construct an $n \times n$ adjacency matrix, where $n$ is the total number of neurons in the MLP.}
\label{fig:adj}
\end{figure}


\subsection{Connectivity Analysis}
\label{sec:connectivity}

To determine the overall connectivity of the pruned network with respect to the reference network, we provide a connectivity analysis based on the heat diffusion signatures of the pruned network. Rather than establishing a thumb rule for estimating the relative connectivity of the pruned network, we aim to provide a novel framework that considers the concept of heat diffusion to gauge network connectivity.    

{\bf Heat Diffusion Signature:} Based on the premise that every network is essentially an undirected graph, the unnormalized graph Laplacian matrix $L$, for a network, is computed from the adjacency matrix using:
\begin{equation}
    L = D - W,
\end{equation}    

\noindent
where \textit{W} is the symmetric adjacency matrix of the graph and D denotes the degree matrix. The Graph Laplacian, $L$ is of the dimension $n\times n$, where $n$ is the total number of nodes/channels in the network. 
We can use the spectrum of Laplacian matrix (i.e., $L=U \Lambda U^T$) compute the heat matrix of the undirected graph~\cite{chung1996lectures} as:
\begin{equation}
H(t) = U e^{-\Lambda t} U^T,
\end{equation}

\noindent
where $H(t)$ and $U$ refer to the heat matrix and the spectral embedding of the Laplacian consisting of $n$ eigenvectors respectively. $\Lambda$ is the diagonal matrix of corresponding eigenvalues, i.e., $\Lambda=Diag[\lambda_1, \hdots, \lambda_n]$.Each $i,j$ element of heat matrix i.e., $H_{ij}(t)$ indicates the amount of heat reaching to node $j$ from node $i$, thereby capturing the scale/time dependent effective connectivity between two nodes. Finally, the heat signature of the network is computed using:
\begin{equation}
S = H(t)A,
\end{equation}
where A is an $n\times1$ binary matrix that assigns each node as a source (1) or sink (0). Here, all the inputs nodes are assigned a value of one (heat source) and the remaining nodes are assigned to be zero (heat sink). Hence, the final matrix S, is an $n\times1$ matrix which gives an estimate of the heat signature of the network/graph at time \textit{t}. 

\noindent
\textbf{Visualizing Adjacency matrices of pruned networks:} Fig.~\ref{fig:adj} shows adjacency matrices for a Multi-Layer Perceptron (MLP), comprising five fully-connected layers parameterized by $\{W_i \in \mathbb{R}^{n_{i+1} \times n_i}\}$, where $n_i \in \{3072,512,256,128,64\}$, and $W_i$ denotes the weight matrix in the $i^{th}$ layer. The dimension of each adjacency matrix is $n \times n$, where $n$ is the total number of channels/nodes in the network. 
To verify the enhanced connectivity resulting from N2NSkip connections, we compare the heat diffusion signature of the pruned adjacency matrices with the heat diffusion signature of the reference network.

\noindent
{\bf Comparing Diffusion Signatures of Networks:}
 We compute the Frobenius norm of the respective heat diffusion signatures of two networks. Given same initial heat distribution, two graphs/networks are expected to have similar heat diffusion signatures as these signatures encode scale dependent topological characterization of the underlying graphs~\cite{sharma2012representation}. 
 %
To ensure fair comparison, we have assumed large scale heat diffusion (large \textit{t} values).
\begin{equation}
F = \|S_{\text {reference}} - S_{\text {prune}}\|_2.
\label{eq:Frob}
\end{equation}
$S_{\text {reference}}$ refers to the heat diffusion signature of the reference network, and $S_{\text {prune}}$ denotes the heat diffusion signature of the pruned network.
The relative connectivity of the pruned network with respect to the reference network is determined by the Frobenius norm (Eq.~\ref{eq:Frob}) of their respective heat signatures. A large value of $F$ would essentially mean that the effective connectivity of the pruned network deviates significantly from the connectivity of the reference network. 

\begin{figure}[!t]
\begin{center}
\begin{tabular}{cc}
\scalebox{0.8}{
\includegraphics[width=6cm, height=4cm]{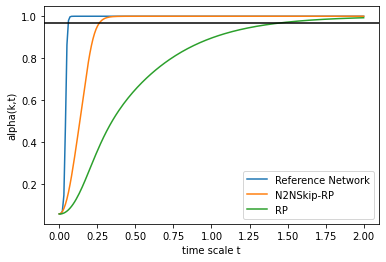}} &
\scalebox{0.8}{
\includegraphics[width=6cm, height=4cm]{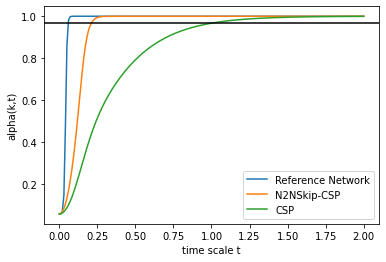}}\\
\scalebox{0.8}{(a) N2NSkip-RP vs RP on VGG19 (Connectivity)} & \scalebox{0.8}{(b) N2NSkip-CSP vs CSP on VGG19 (Connectivity)}\\
\vspace{1mm}
\end{tabular}
\end{center}
\caption{Time taken for heat diffusion signatures of the pruned networks to overlap with that of the baseline network (VGG19). 
Addition of N2NSkip connections leads to significantly faster convergence to a saturated heat distribution ($\alpha(K,t) = 1$), thereby indicating a higher degree of overall connectivity in the pruned network.}
\label{fig:scree}
\end{figure}

\noindent
{\bf Qualitative comparison using Scree Diagram:} 
On an alternate note, given an undirected graph with a heat distribution captured by $H(t)$ at time $t$, the time taken to reach a saturated heat distribution ($\alpha(K,t) = 1$) varies depending on the connectivity of the graph. 
For instance, faster convergence to a saturated heat distribution would directly imply a higher connectivity in the network \cite{teke2017time}.

We use scree diagrams proposed in~\cite{sharma2012representation} to compare the effective connectivity of pruned networks with respect to the reference network. We compute $\alpha(K,t)$, as:
\begin{equation}
\alpha(K,t) = \frac{\sum_{k=2}^{K+1}e^{-t\lambda_k}}{\sum_{k=2}^{n}e^{-t\lambda_k}},
\end{equation}
where $t$ is the time scale parameter which determines the time taken for the heat diffused to attain a saturated heat distribution; $K$ denotes the percentage of total eigenvalues of the Graph Laplacian. 
A plot of $\alpha(K,t)$ for a fixed $K$ depicts rate at which heat diffusion on a graph saturates as $t$ (i.e., diffusion scale) increases. Thus, two graphs/networks that are having similar connectivity are expected to have overlapping curves. Additionally, networks with better connectivity are expected to have their $\alpha$ value saturate faster as compared to other networks with weaker connectivity. We show resultant scree diagrams in Fig.~\ref{fig:scree} where we demonstrate that
 pruned networks with N2NSkip connections attain overlapping (closeby) $\alpha$ curves to that of the reference network. This demonstrates their improved connectivity as compared to pruned sequential networks of the same sparsity.
 

\begin{figure}[!t]
\begin{center}
\begin{tabular}{cccc}
\scalebox{0.8}{
\includegraphics[width=6cm,height=4cm]{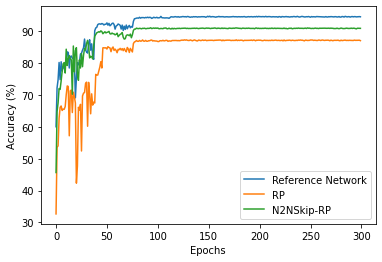}} &
\scalebox{0.8}{
\includegraphics[width=6cm,height=4cm]{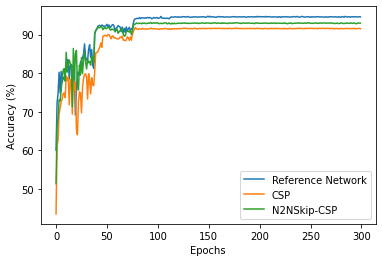}}\\
\scalebox{0.8}{(a) N2NSkip-RP vs RP on ResNet50} & \scalebox{0.8}{(b) N2NSkip-CSP vs CSP on ResNet50}\\
\vspace{1mm}
\end{tabular}
\end{center}
\caption{Superior convergence is observed during training, upon addition of N2NSkip connections to pruned ResNet50 on CIFAR-10 at a compression of 20x. For both RP and CSP, N2NSkip connections enhance overall connectivity, resulting in learning capabilities which are at par with the reference network.}
\label{fig:convergence}
\end{figure}

\section{Experimental Results}
\label{sec:exp}


\textbf{Experimental Setup.}   
All models are trained using SGD with a momemtum of 0.9 and an initial learning rate of 0.05, with a learning rate decay of 0.5 every 30 epochs. We train for 300 epochs on CIFAR-10,CIFAR-100 and Imagenet, with a batch size of 128, and use a weight decay of 0.0005. The accuracies are reported for 5 runs of training and validation. All our evaluation models are trained on four NVIDIA GTX 1080Ti GPUs using PyTorch.

\noindent
\textbf{CIFAR-10 and CIFAR-100.} As shown in Fig. ~\ref{fig:convergence}, the accuracy of N2NSkip networks during the first fifty epochs is nearly equal to the baseline accuracy. This empirically demonstrates that N2NSkip connections improve the overall learning capability of the pruned network.  
Table ~\ref{table:Table1} shows the results on VGG19 and ResNet50 at a density of 10\%, 5\% and 2\%. The addition of N2NSkip connections leads to a significant increase in test accuracy. Additionally, there is a larger increase in accuracy at network densities of 5\% and 2\%, as compared to 10\%. This observation is consistent for both N2NSkip-RP and N2NSkip-CSP, which indicates that N2NSkip connections can be used as a powerful tool to enhance the performance of pruned networks at high compression rates.

\setlength{\tabcolsep}{5pt}
\renewcommand{\arraystretch}{1.2}
\begin{table}
\begin{center}
\label{table:Table1}

\scalebox{0.70}{
\begin{tabular}{cccccccc}
\hline\noalign{\smallskip}
\multirow{2}{*}{{Model}} &  \multirow{2}{*}{{Method}} &  & CIFAR-10 & & & CIFAR-100 &\\
\cline{3-8}
& & 10\% & 5\% & 2\% & 10\% & 5\% & 2\%\\
\hline

\multirow{5}{*}{{VGG19}} & {{Baseline}} & $93.16\pm 0.12$ & - & - & $74.09 \pm 0.15$ & - & -\\
\cline{2-8}
 & RP & $92.08 \pm 0.36$ & $89.43 \pm 0.75$ & $86.52 \pm 1.75$ & $71.23 \pm 0.26$ & $69.82 \pm 0.65$ & $55.43 \pm 1.94$\\
& \textbf{N2NSkip-RP} & $\boldsymbol{92.92 \pm 0.19}$ & $\boldsymbol{92.65 \pm 0.25}$ & $\boldsymbol{91.12 \pm 0.36}$ & $\boldsymbol{72.67 \pm 0.23}$ & $\boldsymbol{72.13 \pm 0.31}$ & $\boldsymbol{61.21 \pm 0.42}$\\
\cline{2-8}
{(143M)} & CSP & $92.79 \pm 0.23$ & $92.14 \pm 0.47$ & $90.35 \pm 0.98$ & $72.83 \pm 0.27$ & $71.92 \pm 0.68$ & $59.92 \pm 1.21$\\
& \textbf{N2NSkip-CSP} & $\boldsymbol{93.02 \pm 0.13}$ & $\boldsymbol{92.86 \pm 0.19}$ & $\boldsymbol{92.12 \pm 0.29}$ & $\boldsymbol{73.72 \pm 0.16}$ & $\boldsymbol{73.05 \pm 0.25}$ & $\boldsymbol{65.45 \pm 0.41}$\\

\hline
\multirow{5}{*}{{ResNet50}} & {{Baseline}} & $95.33 \pm 0.11$ & - & - & $74.94 \pm 0.13$ & - & -\\
\cline{2-8}
& RP  & $88.53 \pm 0.21$ & $86.17 \pm 0.39$ & $83.33 \pm 0.93$ & $67.72 \pm 0.25$ & $62.28 \pm 0.42$ & $51.11 \pm 1.01$\\
& \textbf{N2NSkip-RP} & $\boldsymbol{91.59 \pm 0.16}$ & $\boldsymbol{89.14 \pm 0.24}$ & $\boldsymbol{87.67 \pm 0.51}$ & $\boldsymbol{70.45 \pm 0.14}$ & $\boldsymbol{67.56 \pm 0.28}$ & $\boldsymbol{60.19 \pm 0.51}$\\
\cline{2-8}
{(23M)} & CSP & $93.15 \pm 0.15$ & $92.25 \pm 0.27$ & $89.12 \pm 0.36$ & $69.29 \pm 0.22$ & $65.73 \pm 0.34$ & $55.02 \pm 0.79$\\
& \textbf{N2NSkip-CSP} & $\boldsymbol{94.37 \pm 0.12}$ & $\boldsymbol{93.59 \pm 0.21}$ & $\boldsymbol{92.26 \pm 0.31}$ & $\boldsymbol{72.37 \pm 0.15}$ & $\boldsymbol{70.43 \pm 0.27}$ & $\boldsymbol{63.16 \pm 0.48}$\\

\hline
\end{tabular}}
\end{center}
\caption{Test Accuracy of pruned ResNet50 and VGG19 on CIFAR-10 and CIFAR-100 with either RP or CSP as the preliminary pruning step. M denotes total number of parameters in millions.}

\label{table:Table1}
\end{table}

\noindent
\textbf{ImageNet LSVRC 2012.} Table~\ref{table:Table2} shows the performance of N2NSkip connections on ImageNet at compression rates of 2x, 3.3x and 5x (density of 50\%, 30\% and 20\%). We have used ResNet50 to demonstrate the enhanced performance resulting from N2NSkip connections. Similar to the results reported in Table ~\ref{table:Table1}, networks with N2NSkip connections significantly outperform purely sequential networks (RP and CSP), especially at higher sparsity levels. 

\setlength{\tabcolsep}{2pt}
\begin{table}[t]
\begin{center}
\begin{tabular}{cc}
\scalebox{0.66}{
\begin{tabular}{ccccc}
\hline\noalign{\smallskip}
\multirow{2}{*}{Model} & \multirow{2}{*}{Method} & & Density &  \\ 
\cline{3-5}
& & 50\% & 30\% & 20\% \\
\hline
\multirow{3}{*}{ResNet50} & {Baseline} & $74.70 \pm 0.26$ & - & -\\
 & {CSP} & $73.42 \pm 0.29$ &  $70.42 \pm 0.37$ & $68.67 \pm 0.65$\\
 (23M) & {\textbf{N2NSkip-CSP}} & $\boldsymbol{74.59 \pm 0.22}$ & $\boldsymbol{72.89 \pm 0.33}$ & $\boldsymbol{72.09 \pm 0.45}$\\
\hline
\end{tabular}}
&
\scalebox{0.66}{
\begin{tabular}{ccccc}
\hline\noalign{\smallskip}
\multirow{2}{*}{Model} & \multirow{2}{*}{Method} & & Density &  \\ 
\cline{3-5}
& & 50\% & 30\% & 20\% \\
\hline
\multirow{3}{*}{ResNet50} & {Baseline} & $74.70\pm 0.28$ & - & -\\
 & {RP} & $72.46 \pm 0.32$ & $68.65 \pm 0.45$ & $65.32 \pm 0.97$\\
 (23M) & \textbf{N2NSkip-RP} & $\boldsymbol{74.12 \pm 0.29}$ & $\boldsymbol{71.19 \pm 0.39}$ & $\boldsymbol{70.03 \pm 0.51}$\\

\hline
\end{tabular}}
\end{tabular}
\end{center}
\caption{Test Accuracy of pruned ResNet50 on ImageNet with either CSP (left) or RP (right) as the preliminary pruning step. }
\label{table:Table2}
\end{table}

\setlength{\tabcolsep}{4pt}
\renewcommand{\arraystretch}{1.2}
\begin{table}[!t]
\begin{center}
\scalebox{0.7}{

\begin{tabulary}{\linewidth}{ccccccc}
\hline
\multirow{2}{*}{Model} & \multirow{2}{*}{Method} &  \multicolumn{4}{c}{Density}\\ 
\cline{3-6}
& & 50\% & 10\% & 5\% & 2\%\\
\hline
\multirow{4}{*}{\textbf{VGG19}} & RP & $3.6\times10^{-3}$ & $4.2\times10^{-1}$ & $2.3\times10^{-1}$ & $6.5\times10^{0}$\\
& \textbf{N2NSkip-RP} &  \textbf{\boldmath$2.8\times10^{-6}$} & \textbf{\boldmath$4.9\times10^{-5}$} & \textbf{\boldmath$9.9\times10^{-4}$} & \textbf{\boldmath$1.3\times10^{-3}$}\\
\cline{2-6}

& CSP & $7.9\times10^{-3}$ & $7.1\times10^{-5}$ & $9.1\times10^{-2}$ & $2.3\times10^{0}$\\
& \textbf{N2NSkip-CSP} &  \textbf{\boldmath$1.4\times10^{-6}$} & \textbf{\boldmath$2.5\times10^{-5}$} & \textbf{\boldmath$6.2\times10^{-5}$} & \textbf{\boldmath$3.3\times10^{-4}$}\\
\hline

\hline
\multirow{4}{*}{\textbf{ResNet50}} & RP & $4.4\times10^{-3}$ & $3.9\times10^{-2}$ & $4.5\times10^{-1}$ & $1.2\times10^{1}$\\
& \textbf{N2NSkip-RP} & \textbf{\boldmath$8.1\times10^{-6}$} & \textbf{\boldmath$5.5\times10^{-5}$} & \textbf{\boldmath$3.8\times10^{-4}$} & \textbf{\boldmath$5.6\times10^{-3}$}\\
\cline{2-6}

& {CSP} & $7.9\times10^{-3}$ & $6.7\times10^{-5}$ & $6.1\times10^{-2}$ & $9.2\times10^{0}$\\
& \textbf{N2NSkip-CSP} &  \textbf{\boldmath$5.3\times10^{-6}$} & \textbf{\boldmath$1.7\times10^{-5}$} & \textbf{\boldmath$4.2\times10^{-5}$} & \textbf{\boldmath$8.9\times10^{-4}$}\\

\hline

\end{tabulary}}
\end{center}
\caption{Difference in connectivity of pruned models with respect to the reference network at saturated heat distribution. The difference is minimum for N2NSkip networks, thereby indicating superior overall connectivity in the model.}
\label{table:Table3}
\end{table}
\setlength{\tabcolsep}{1.4pt}

\noindent
\textbf{Overall Connectivity.} The deviation in connectivity between the pruned and reference networks, given by the Frobenius norm between their respective heat diffusion signatures, is reported in Table~\ref{table:Table3}. We obtain the respective diffusion signatures from their corresponding weighted adjacency matrices (similar to Fig. ~\ref{fig:adj}).  
To establish that N2NSkip connections cause an appreciable improvement in connectivity, we compare the values of $F$ for pruned networks with and without N2NSkip connections. 
The Frobenius norm for pruned networks with N2NSkip connections is considerably lesser than the Frobenius norm for pruned networks without N2NSkip connections (CSP and RP). 

\subsection{N2NSkip vs conventional Skip connections}
\label{sec:resskip}
\begin{figure}[h]
\begin{center}
\begin{tabular}{cc}
\scalebox{0.9}{\includegraphics[width=5.4cm, height=3.6cm]{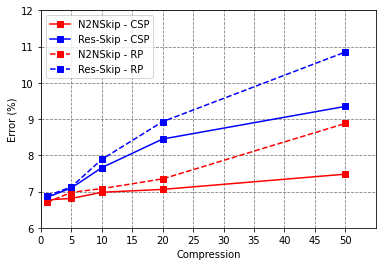}} & 
\scalebox{0.9}{\includegraphics[width=5.4cm, height=3.6cm]{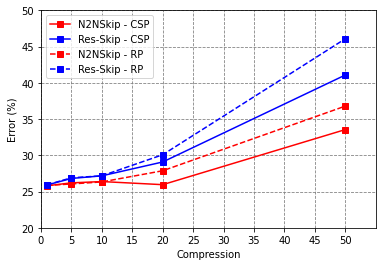}}\\
\scalebox{0.8}{(a) N2NSkip vs ResSkip on VGG19} & \scalebox{0.8}{(b) N2NSkip vs ResSkip on VGG19}\\
\scalebox{0.8}{(CIFAR-10)} & \scalebox{0.8}{(CIFAR-100)} \\
\vspace{1mm}
\end{tabular}
\end{center}
\caption{N2NSkip vs ResSkip for pruning VGG19 on CIFAR-10 and CIFAR-100.  At higher sparsity levels, N2NSkip connections result in lower performance degradation as compared to ResSkip connections.}
\label{fig:VGG19Comp}
\end{figure}

In order to demonstrate the superiority of N2NSkip connections over conventional skip connections (ResSkip for brevity), we  compared the effect of adding N2NSkip (red plots) vs ResSkip (blue plots) connections to VGG19 while maintaining the same sparsity. 
Fig.~\ref{fig:VGG19Comp} shows the test error for (N2NSkip + VGG-19) vs (ResSkip + VGG) on CIFAR-10 and CIFAR-100. N2NSkip connections consistently produce significantly lower test error, especially at higher compression rates of 20x and 50x. Here, ResSkip-RP and ResSkip-CSP refer to the addition of ResSkip connections to VGG19 after randomized and connection sensitivity pruning respectively.

\section{Conclusion}
We proposed neuron-to-neuron skip (N2NSkip) connections which act as sparse weighted skip connections between sequential layers of the network. While maintaining the same density, we found that adding N2NSkip connections to a pruned network results in significantly superior accuracy, convergence and connectivity. Additionally, we also provide a new approach to analyze the connectivity of neural networks using heat diffusion, thereby providing a different perspective on evaluating the efficacy of network architectures. We demonstrated that pruned networks with N2NSkip connections have minimal loss in connectivity with respect to the reference network. We believe that the field of deep learning can benefit greatly from similar explorations in graph theory. 

    

\bibliography{egbib}
\end{document}